\documentclass[runningheads]{llncs}
\usepackage{graphicx}
\usepackage{subcaption}
\usepackage{tabularx}
\newcolumntype{C}[1]{>{\centering\arraybackslash}p{#1}}
\usepackage{amsmath,amssymb} 
\usepackage{color}

\usepackage{tabu}
\usepackage{multirow}
\usepackage{float}

\newcommand{\minisection}[1]{\vspace{0.04in} \noindent {\bf #1}\ \ }

\begin{document}
\pagestyle{headings}
\mainmatter
\def\ECCV18SubNumber{2528}  
\title{Transferring GANs: generating images from limited data}


\titlerunning{Transferring GANs: generating images from limited data}
\authorrunning{Wang et al.}


\author{Yaxing Wang, Chenshen Wu, Luis Herranz, Joost van de Weijer, \\
Abel Gonzalez-Garcia, Bogdan Raducanu \\
{\small\{yaxing, chenshen, lherranz, joost, agonzgarc, bogdan\}@cvc.uab.es}
}
\institute{Computer Vision Center\\Universitat Aut\`{o}noma de Barcelona, Spain}

\maketitle
\iftrue
\begin{abstract}
Transferring knowledge of pre-trained networks to new domains by means of fine-tuning is a widely used practice for applications based on discriminative models. To the best of our knowledge this practice has not been studied within the context of generative deep networks. Therefore, we study domain adaptation applied to image generation with generative adversarial networks. We evaluate several aspects of domain adaptation, including the impact of target domain size, the relative distance between source and target domain, and the initialization of conditional GANs. Our results show that using knowledge from pre-trained networks can shorten the convergence time and can significantly improve the quality of the generated images, especially when target data is limited. We show that these conclusions can also be drawn for conditional GANs even when the pre-trained model was trained without conditioning. Our results also suggest that density is more important than diversity and a dataset with one or few densely sampled classes is a better source model than more diverse datasets such as ImageNet or Places.

\end{abstract}

\section{Introduction} 


Generative Adversarial Networks (GANs) can generate samples from complex image distributions~\cite{goodfellow2014generative}. They consist of two networks: a discriminator which aims to separate real images from fake (or generated) images, and a generator which is simultaneously optimized to generate images which are classified as real by the discriminator. The theory was later extended to the case of conditional GANs where the generative process is constrained using a conditioning prior~\cite{mirza2014conditional}  which is provided as an additional input. GANs have further been widely applied in applications, including super-resolution~\cite{ledig2016photo}, 3D object generation and reconstruction~\cite{smith2017improved}, human pose estimation~\cite{ma2017pose}, and
age estimation~\cite{zhang2017age}.

Deep neural networks have obtained excellent results for discriminative classification problems for which large datasets exist; for example on the ImageNet dataset which consists of over 1M images~\cite{krizhevsky2012imagenet}. However, for many problems the amount of labeled data is not sufficient to train the millions of parameters typically present in these networks. Fortunately, it was found that the knowledge contained in a network trained on a large dataset (such as ImageNet) can easily be transferred to other computer vision tasks. Either by using these networks as off-the-shelf feature extractors~\cite{azizpour2016factors}, or by adapting them to a new domain by a process called fine tuning~\cite{oquab2014learning}. In the latter case, the pre-trained network is used to initialize the weights for a new task (effectively transferring the knowledge learned from the source domain), which are then fine tuned with the training images from the new domain. It has been shown that much fewer images were required to train networks which were initialized with a pre-trained network.

%
GANs are in general trained from scratch. The procedure of using a pre-trained network for initialization -- which is very popular for discriminative networks -- is to the best of our knowledge not used for GANs. However, like in the case of discriminative networks, the number of parameters in a GAN is vast; for example the popular DC-GAN architecture~\cite{radford2015unsupervised} requires 36M parameters to generate an image of 64x64. Especially in the case of  domains which lack many training images, the usage of pre-trained GANs could significantly improve the quality of the generated images.

Therefore, in this paper, we set out to evaluate the usage of pre-trained networks for GANs. The paper has the following contributions:
\begin{enumerate}
\item We evaluate several transfer configurations, and show that pre-trained networks can effectively accelerate the learning process and provide useful prior knowledge when data is limited.
\item We study how the relation between source and target domains impacts the results, and discuss the problem of choosing a suitable pre-trained model, which seems more difficult than in the case of discriminative tasks.
\item We evaluate the transfer from unconditional GANs to conditional GANs for two commonly used methods to condition GANs. 
\end{enumerate}






\section{Related Work}

\minisection{Transfer learning/domain transfer:}
Learning how to transfer knowledge from a source domain to target domain is a well studied problem in computer vision~\cite{pan2010survey}. In the deep learning era, complex knowledge is extracted during the training stage on large datasets~\cite{russakovsky2015imagenet,zhou2014learning}.  Domain adaptation by means of fine tuning a pre-trained network has become the default approach for many applications with limited training data or slow convergence~\cite{donahue2014decaf,oquab2014learning}. 

Several works have investigated transferring knowledge to unsupervised or sparsely labeled domains. Tzeng et al.~\cite{tzeng2015simultaneous}  optimized for domain invariance, while transferring task information that is present in the correlation between the classes of the source domain. Ganin et al.~\cite{ganin2016domain} proposed to learn domain invariant features by means of a gradient reversal layer. A network simultaneously trained on these invariant features can be transfered to the target domain. Finally, domain transfer has also been studied for networks that learn metrics~\cite{hu2015deep}.  In contrast to these methods, we do not focus on transferring discriminative features, but transferring knowledge for image generation.


\minisection{GAN:}
Goodfellow et al.~\cite{goodfellow2014generative} introduced the first GAN model for image generation. Their architecture uses a series of fully connected layers and thus is limited to simple datasets. 
When approaching the generation of real images of higher complexity, convolutional architectures have shown to be a more suitable option. 
%
%
Shortly afterwards, Deep Convolutional GANs (DC-GAN)  quickly became the standard GAN architecture for image generation problems~\cite{radford2015unsupervised}.
In DC-GAN, the generator sequentially up-samples the input features by using fractionally-strided convolutions, whereas the discriminator uses normal convolutions to classify the input images. Recent multi-scale architectures \cite{denton2015deep,huang2017stacked,karras2017progressive} can effectively generate high resolution images. It was also found that ensembles can be used to improve the quality of the generated distribution~\cite{wang2016ensembles}. 



Independently of the type of architecture used, GANs present multiple challenges regarding their training, such as convergence properties, stability issues, or mode collapse.  
%
%
Arjovksy et al.~\cite{arjovsky2017towards} showed that the original GAN loss~\cite{goodfellow2014generative} are unable to properly deal with ill-suited distributions such as those with disjoint supports, often found during GAN training.
%
Addressing these limitations the Wassertein GAN \cite{arjovsky2017wasserstein} uses the Wasserstein distance as a robust loss, yet requiring the generator to be 1-Lipschitz. This constrain is originally enforced by clipping the weights. Alternatively, an even more stable solution is adding a gradient penalty term to the loss (known as WGAN-GP)~\cite{gulrajani2017improved}. 
%



\minisection{cGAN:} 
Conditional GANs (cGANs) \cite{mirza2014conditional} are a class of GANs that use a particular attribute as a prior to build conditional generative models. 
%
Examples of conditions are class labels~\cite{odena2016conditional,perarnau2016invertible,grinblat2017class}, text~\cite{reed2016text2image,zhang2017text2image}, another image (image translation~\cite{kim2017image2image,zhu2017image2image} and style transfer~\cite{dumoulin2017artistic}).

Most cGAN models~\cite{mirza2014conditional,zhang2017text2image,dumoulin2017adversarially,sricharan2017semisuper} apply their condition in both generator and discriminator by concatenating it to the input of the layers, i.e. the noise vector for the first layer or the learned features for the internal layers.  Instead, in \cite{dumoulin2017artistic}, they include the conditioning in the batch normalization layer.
%
The AC-GAN framework~\cite{odena2016conditional} extends the discriminator with an auxiliary decoder to reconstruct class-conditional information.
Similarly, InfoGAN~\cite{chen2016infogan} reconstructs a subset of the latent variables from which the samples were generated. 
Miyato et al.~\cite{miyato2018projection} propose another modification of the discriminator
based on a ‘projection layer’ that uses the inner product between the conditional information and the intermediate output to compute its loss. 

\section{Generative Adversarial Networks}

\subsection{Loss functions}
\label{sec:GAN_basics}
A GAN consists of a generator $G$ and a discriminator $D$~\cite{goodfellow2014generative}. The aim is to train a generator $G$ which generates samples that are indistinguishable from the real data distribution. The discriminator is optimized to distinguish samples from the real data distribution $p_{data}$ from those of the fake (generated) data distribution $p_g$. The generator takes noise $z \sim p_z$ as input, and generates samples $G\left(z\right)$ with a distribution $p_g$. The networks are trained with an adversarial objective.  The generator is optimized to generate samples which would be classified by the discriminator as belonging to the real data distribution. The minimax game objective is given by:
\begin{gather}
G^*  = \mathop \textnormal{argmin}\limits_G \mathop {\max }\limits_D \mathcal{L}_{GAN}\left(G,D\right) \\
\mathcal{L}_{GAN}\left(G,D\right) = \mathbb{E}_{x\sim p_{data}} [\log D(x)] + \mathbb{E}_{z\sim p_z}[ \log(1-D(G(z)))]
\label{eq:GAN}
\end{gather}

In the case of WGAN-GP \cite{gulrajani2017improved} the two loss functions are:
\begin{equation}
\begin{aligned}
& \mathcal{L}_{WGAN-GP}\left(D\right) = -\mathbb{E}_{x\sim p_{data}} [D(x)] + \mathbb{E}_{z\sim p_z}[ D(G(z))]\\ & + \lambda \mathbb{E}_{x\sim p_{data},{z\sim p_z},\alpha\sim\left(0,1\right)}\left[\left(\Vert \nabla D\left(\alpha x+ \left(1-\alpha \right)G(z) \right)\Vert_2 -1 \right)^2 \right]
\label{eq:WGANGP_D}
\end{aligned}
\end{equation}
\begin{equation}
\mathcal{L}_{WGAN-GP}\left(G\right) = -\mathbb{E}_{z\sim p_z}[ D(G(z))]
\label{eq:WGANGP_G}
\end{equation}



\subsection{Evaluation Metrics}
 

Evaluating GANs is notoriously difficult \cite{theis2015note} and there is no clear agreed reference metric yet. In general, a good metric should measure the quality and the diversity in the generated data. Likelihood has been shown to not correlate well with these requirements~\cite{theis2015note}. Better correlation with human perception has been found in the widely used Inception Score~\cite{salimans2016improved}, but recent works have also shown its limitations~\cite{zhou2018activation}. In our experiments we use two recent metrics that show better correlation in recent studies \cite{im2018quantitatively,borji2018pros}. While not perfect, we believe they are satisfactory enough to help us to compare the models in our experiments.

\subsubsection{Fr\'echet Inception Distance~\cite{heusel2017gans}} 
The similarity between two sets is measured as their Fr\'echet distance (also known as Wasserstein-2 distance) in an embedded space. The embedding is computed using a fixed convolutional network (an Inception model) up to a specific layer. The embedded data is assumed to follow a multivariate normal distribution, which is estimated by computing their mean and covariance. In particular, the FID is computed as
\begin{equation}
\textnormal{FID}\left ( \mathcal{X}_{1},\mathcal{X}_{2} \right )=\left \| \mu_{1}-\mu_2  \right \|^2_2 + \operatorname{Tr} \left ( \Sigma_{1} + \Sigma_2 -2\left ( \Sigma_{1}\Sigma_2 \right ) ^\frac{1}{2}\right )
\end{equation}
Typically, $\mathcal{X}_1$ is the full dataset with real images, while $\mathcal{X}_2$ is a set of generated samples. We use FID as our primary metric, since it is efficient to compute and correlates well with human perception \cite{heusel2017gans}.

\subsubsection{Independent Wasserstein (IW) critic~\cite{danihelka2017comparison}}
This metric uses an independent critic $\hat{D}$ only for evaluation. This independent critic will approximate the Wasserstein distance \cite{arjovsky2017towards} between two datasets $\mathcal{X}_{1}$ and $\mathcal{X}_{2}$ as
\begin{equation}
\textnormal{IW}\left ( \mathcal{X}_{1},\mathcal{X}_{2} \right )=\mathbb{E}_{x\sim \mathcal{X}_{1}}\left( \hat{D}\left( x \right) \right) - \mathbb{E}_{x\sim \mathcal{X}_{2}}\left( \hat{D}\left( x \right) \right)
\end{equation}
In this case, $\mathcal{X}_1$ is typically a validation set, used to train the independent critic. We report IW only in some experiments, due to the larger computational cost that requires training a network for each measurement.

\setlength{\tabcolsep}{4pt}
\begin{table}[tb]
	\begin{center}
		\caption{FID/IW (the lower the better / the higher the better) for different transfer configurations. ImageNet was used as source dataset and LSUN Bedrooms as target (100K images).}
		\label{table:transfer_configuration}
\begin{tabular}{lllll}
\hline
\multicolumn{1}{c}{Generator}     & \multicolumn{2}{c}{Scratch} & \multicolumn{2}{c}{Pre-trained} \\
\multicolumn{1}{c}{Discriminator} & \multicolumn{1}{c}{Scratch} & \multicolumn{1}{c}{Pre-trained} & \multicolumn{1}{c}{Scratch} & \multicolumn{1}{c}{Pre-trained} \\
\hline
$\textnormal{FID}\left(\mathcal{X}^{tgt}_{data},\mathcal{X}^{tgt}_{gen}\right)$ & 32.87 & 30.57 & 56.16 & \textbf{24.35}              \\
$\textnormal{IW}\left(\mathcal{X}^{tgt}_{val},\mathcal{X}^{tgt}_{gen}\right)$ & -4.27 & -4.02 & -6.35 & \textbf{-3.88} \\
\hline
\end{tabular}
	\end{center}
\end{table}
\setlength{\tabcolsep}{1.4pt}

\section{Transferring GAN representations}

\subsection{GAN adaptation}\label{GAN_domain_adaptation}
To study the effect of domain transfer for GANs we will use the WGAN-GP~\cite{gulrajani2017improved}  architecture which uses ResNet in both generator and discriminator. 
This architecture has been experimentally demonstrated to be stable and robust against mode collapse~\cite{gulrajani2017improved}.
The generator consists of one fully connected layer, four Residual Blocks and one convolution layer, and the Discriminator has same setting. The same architecture is used for conditional GAN.

\subsubsection{Implementation details}
We generate images of 64$\times$64 pixels, using standard values for hyperparameters. The source models\footnote{The pretrained models are available at https://github.com/yaxingwang/Transferring-GANs.} are trained with a batch of 128 images during 50K iterations (except 10K iterations for CelebA) using Adam \cite{kingma2014adam} and a learning rate of 1e-4. For fine tuning we use a batch size of 64 and a learning rate of 1e-4 (except 1e-5 for 1K target samples). Batch normalization and layer normalization are used in the generator and discriminator respectively.
 

\subsection{Generator/discriminator transfer configuration}
\label{sec:transferConf}
The two networks of the GAN (generator and discriminator) can be initialized with either random or pre-trained weights (from the source networks). In a first experiment we consider the four possible combinations using a GAN pre-trained with ImageNet and 100K samples of LSUN bedrooms as target dataset. The source GAN was trained for 50K iterations. The target GAN was trained for (additional) 40K iterations.

Table~\ref{table:transfer_configuration} shows the results. Interestingly,  we found that transferring the discriminator is more critical than transferring the generator. The former helps to improve the results in both FID and IW metrics, while the latter only helps if the discriminator was already transferred, otherwise harming the performance. Transferring both obtains the best result. We also found that training is more stable in this setting. Therefore, in the rest of the experiments we evaluated either training both networks from scratch or pre-training both (henceforth simply referred to as \textit{pre-trained}).

\begin{figure}[t]
    \centering
   
   \begin{subfigure}{0.3\textheight}
   \centering
   \includegraphics[width=\textwidth]{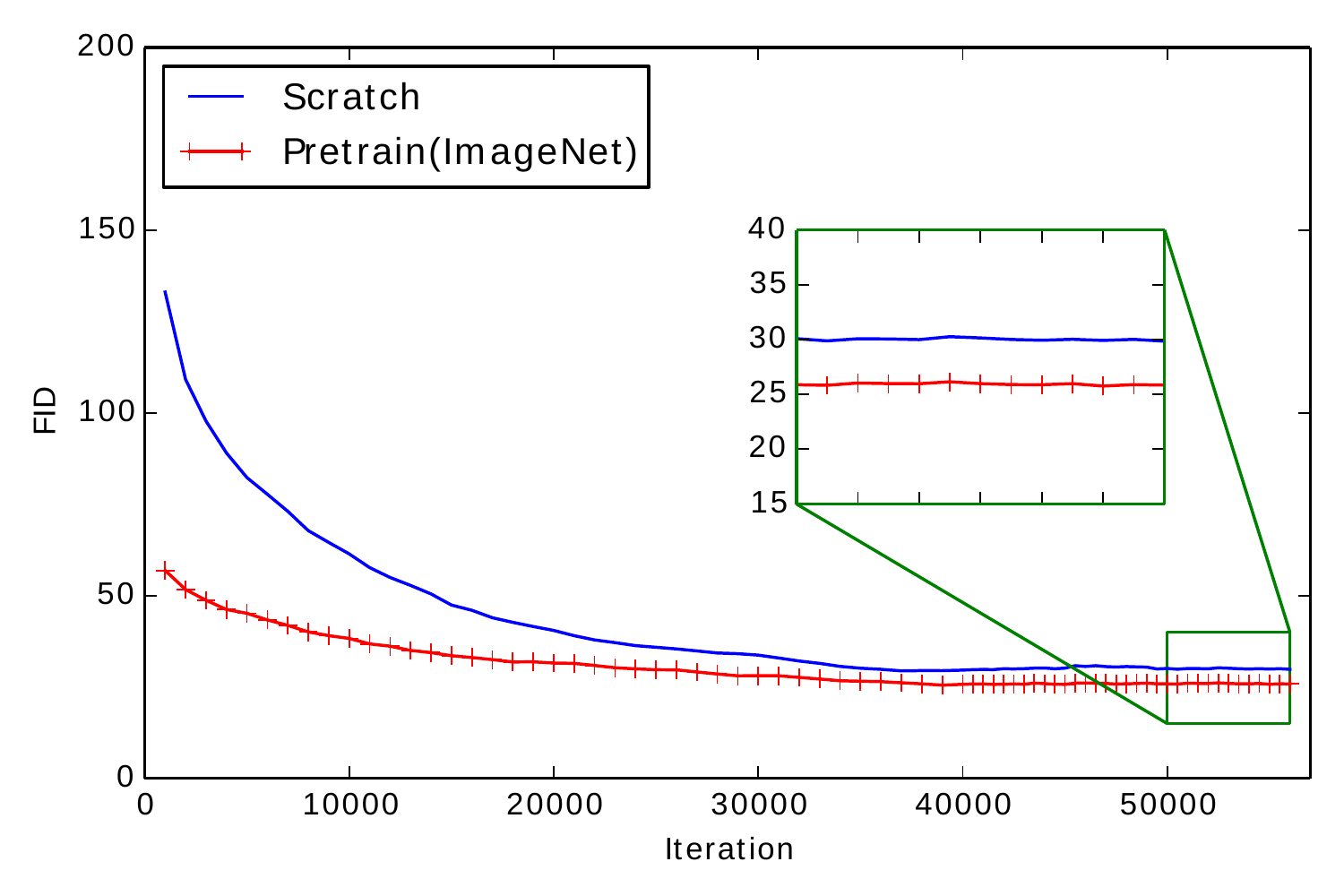}
   \caption{Unconditional GAN(FID)}
   \end{subfigure}%
   \hspace{2mm}
   \begin{subfigure}{0.29\textheight}
   \centering
   \includegraphics[width=\textwidth]{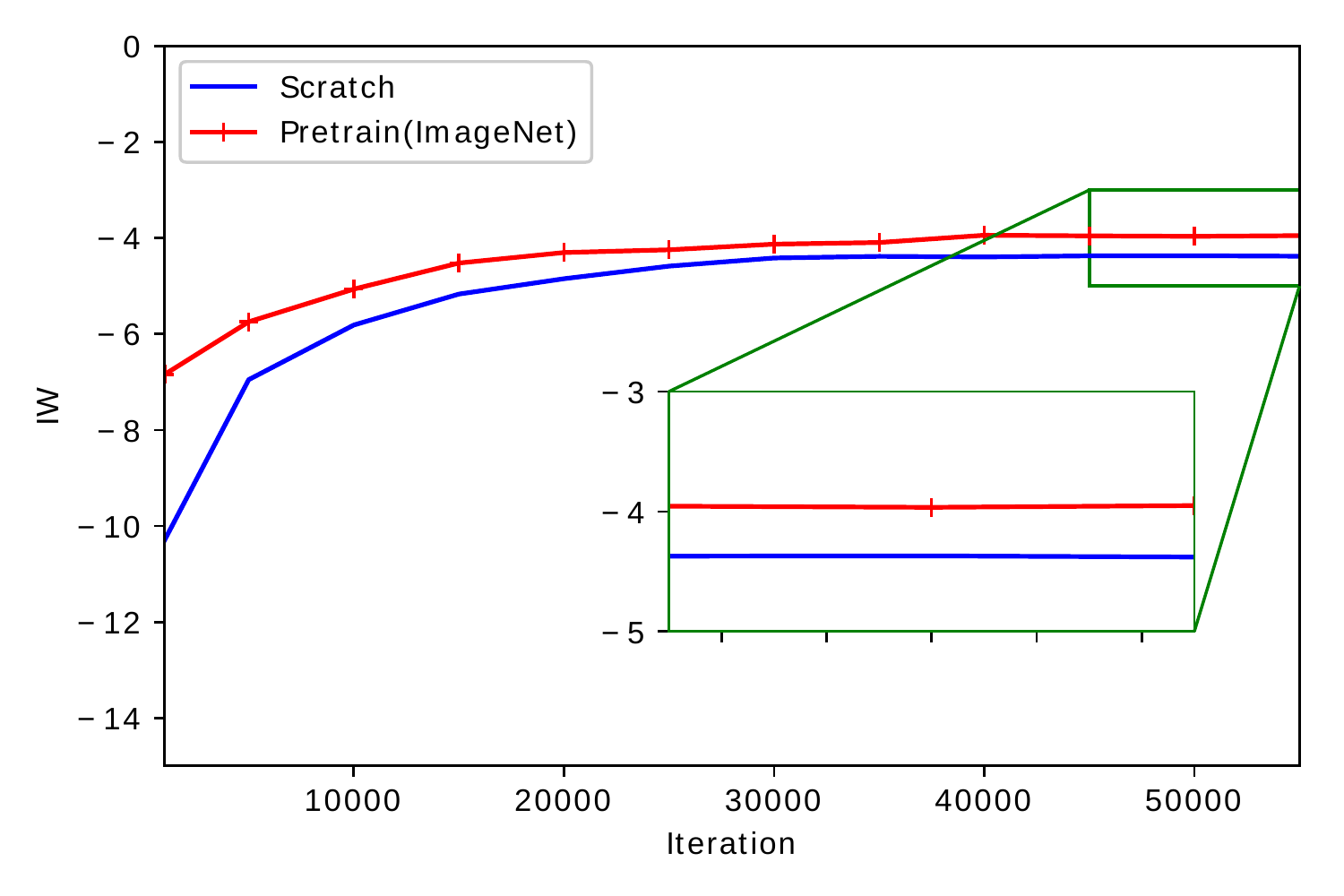}
   \caption{Unconditional GAN(IW)}

   \end{subfigure}
   \caption{Evolution of evaluation metrics when trained from scratch or using a pre-trained model for unconditional GAN measured with (a) FID and (b) IW (source: ImageNet, target: LSUN Bedrooms, metrics: FID and IW). The curves are smoothed for easier visualization by averaging in a window of a few iterations.
    }   \label{fig:fid_acgan}
\end{figure}

Figure~\ref{fig:fid_acgan} shows the evolution of FID and IW during the training process with and without transfer. 
Networks adapted from a pre-trained model can generate images of given scores in significantly fewer iterations. Training from scratch for a long time manages to reduce this gap significantly, but pre-trained GANs can generate images with good quality already with much fewer iterations. Figures~\ref{fig:target_size} and \ref{fig:progression_sources} show specific examples illustrating visually these conclusions.

\subsection{Size of the target dataset}
\label{sec:sizeTarget}

The number of training images is critical to obtain realistic images, in particular as the resolution increases. Our experimental settings involve generating images of 64$\times$64 pixels, where GANs typically require hundreds of thousands of training images to obtain convincing results. We evaluate our approach in a challenging setting where we use as few as 1000 images from the LSUN Bedrooms dataset, and using ImageNet as source dataset. Note that, in general, GANs evaluated on LSUN Bedrooms use the full set of 3M million images.


\setlength{\tabcolsep}{4pt}

\begin{table}[tb]
	\begin{center}
		\caption{FID/IW for different sizes of the target set (LSUN Bedrooms) using ImageNet as source dataset.}
		\label{table:transfer_size}
        \resizebox{\textwidth}{!}{
		\begin{tabular}{cccccccc}
             Target samples & 1K & 5K & 10K & 50K & 100K & 500K & 1M \\
             \hline
             From scratch & 256.1/-33.3 & 86.0/-18.5 & 73.7/-15.3 & 45.5/-7.4 & 32.9/-4.3 & 24.9/-3.6 & 21.0/-2.9\\
             Pre-trained & 93.4/-22.5 & 74.3/-16.3 & 47.0/-7.0 & 29.6/-4.56 & 24.4/-4.0 & 21.6/-3.2 & 18.5/-2.8 \\
			\hline
		\end{tabular}
        }
	\end{center}
\end{table}

Table~\ref{fig:target_size} shows FID and IW measured for different amounts of training samples of the target domain. As the training data becomes scarce, the training set implicitly becomes less representative of the full dataset (i.e. less diverse). In this experiment, a GAN adapted from the pre-trained model requires roughly between two and five times fewer images to obtain a similar score than a GAN trained from scratch. FID and IW are sensitive to this factor, so in order to have a lower bound we also measured the FID between the specific subset used as training data and the full dataset. With 1K images this value is even higher than the value for generated samples after training with 100K and 1M images.

Intializing with the pre-trained GAN helps to improve the results in all cases, being more significant as the target data is more limited. The difference with the lower bound is still large, which suggests that there is still field for improvement in settings with limited data. 

Figure~\ref{fig:target_size} shows images generated at different iterations. As in the previous case, pre-trained networks can generate high quality images already in earlier iterations, in particular with sharper and more defined shapes and more realistic fine details. Visually, the difference is also more evident with limited data, where learning to generate fine details is difficult, so adapting pre-trained networks can transfer relevant prior information.

\begin{figure}[t]
	\centering
    \begin{tabular}{ccc}
	\rotatebox{90}{\begin{tabular}{ C{1.45cm} C{1.45cm} C{1.45cm} C{1.45cm}}1M & 100K & 10K & 1K\end{tabular}} &
    \includegraphics[width=0.4\textwidth]{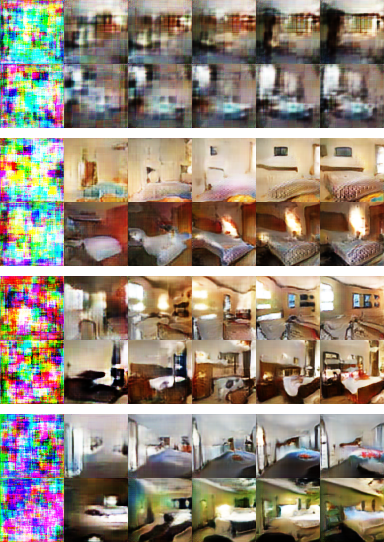} & \includegraphics[width=0.4\textwidth]{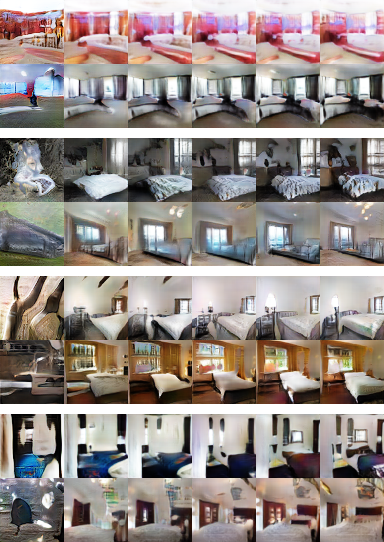}
    \tabularnewline
& From scratch &
Pre-trained (ImageNet)
    \end{tabular}
	\caption{Images generated at different iterations (from 0 and 10000, step 2000) for LSUN bedrooms training from scratch and from a pre-trained network. Better viewed in electronic version.}
	\label{fig:target_size}
\end{figure}

\subsection{Source and target domains}

\setlength{\tabcolsep}{4pt}
\begin{table}[tb]
	\begin{center}
		\caption{Datasets used in the experiments.}
		\label{table:datasets}
		\begin{tabular}{ccccc}
			\hline\noalign{\smallskip}
			 Source datasets & ImageNet~\cite{russakovsky2015imagenet} & Places~\cite{zhou2014learning} & Bedrooms~\cite{yu2015construction} &CelebA~\cite{liu2015deep} \\	
			\noalign{\smallskip}
			 Number of images & 1M & 2.4M & 3M & 200K \\
             Number of classes & 1000 & 205 & 1 & 1 \\
             \hline
            Target datasets & Flower~\cite{nilsback2008automated}   & Kitchens~\cite{yu2015construction} & LFW~\cite{huang2007labeled}  & Cityscapes~\cite{cordts2016cityscapes}\\
			 Number of images & 8K &  50K & 13K  & 3.5K \\
             Number of classes & 102 & 1 & 1 & 1 \\
            
			\hline
		\end{tabular}
	\end{center}
\end{table}

\setlength{\tabcolsep}{4pt}
\begin{table}[tb]
	\begin{center}
		\caption{Distance between target real data and target generated data $\textnormal{FID/IW}\left(\mathcal{X}^{tgt}_{data},\mathcal{X}^{tgt}_{gen}\right)$.}
		\label{table:source_target_generated}
		\begin{tabular}{cccccc}
			\hline\noalign{\smallskip}
			 \begin{tabular}[c]{@{}l@{}}Source $\rightarrow$ \\ Target $\downarrow$ \end{tabular} & Scratch & ImageNet  & Places & Bedrooms & CelebA \\
			\noalign{\smallskip}
			\hline
			\noalign{\smallskip}
			Flowers & 71.98/-13.62 & \textbf{54.04}/\textbf{-3.09} & 66.25/-5.97 & 56.12/-5.90 & 67.96/-12.64 \\
            Kitchens & 42.43/-7.79 & 34.35/-4.45 & 34.59/\textbf{-2.92} & \textbf{28.54}/-3.06 & 38.41/-4.98 \\
            LFW & 19.36/-8.62 & 9.65/-5.17 & 15.02/-6.61 & 7.45/-3.61 & \textbf{7.16}/\textbf{-3.45} \\
            Cityscapes & 155.68/-9.32 & \textbf{122.46}/-9.00 & 151.34/-8.94 & 123.21/-8.44 & 130.64/\textbf{-6.40} \\
			\hline
		\end{tabular}
	\end{center}
\end{table}

The domain of the source model and its relation with the target domain are also a critical factor. We evaluate different combinations of source domains and target domains (see Table~\ref{table:datasets} for details). As source datasets we used ImageNet, Places, LSUN Bedrooms and CelebA. Note that both ImageNet and Places cover wide domains, with great diversity in objects and scenes, respectively, while LSUN Bedrooms and CelebA cover more densely a narrow domain. As target we used smaller datasets, including Oxford Flowers, LSUN Kitchens (a subset of 50K out of 2M images), Label Faces in the Wild (LFW) and CityScapes.

We pre-trained GANs for the four source datasets and then trained five GANs for each of the four target datasets (from scratch and initialized with each of the source GANs). The FID and IW after fine tuning  are shown in Table~\ref{table:source_target_generated}. Pre-trained GANs achieve significantly better results. Both metrics generally agree but there are some interesting exceptions. The best source model for Flowers as target is ImageNet, which is not surprising since it contains also flowers, plants and objects in general. It is more surprising that Bedrooms is also competitive according to FID (but not so much according to IW). The most interesting case is perhaps Kitchens, since Places has several thousands of kitchens in the dataset, yet also many more classes that are less related. In contrast, bedrooms and kitchens are not the same class yet still very related visually and structurally, so the much larger set of related images in Bedrooms may be a better choice. Here FID and IW do not agree, with FID clearly favoring Bedrooms, and even the less related ImageNet, over Places, while IW preferring Places by a small margin. 
As expected, CelebA is the best source for LFW, since both contain faces (with different scales though), but Bedroom is surprisingly very close to the performance in both metrics. For Cityscapes all methods have similar results (within a similar range), with both high FID and IW, perhaps due to the large distance to all source domains.

\begin{figure}[t]
    \centering
     \includegraphics[width=\textwidth]{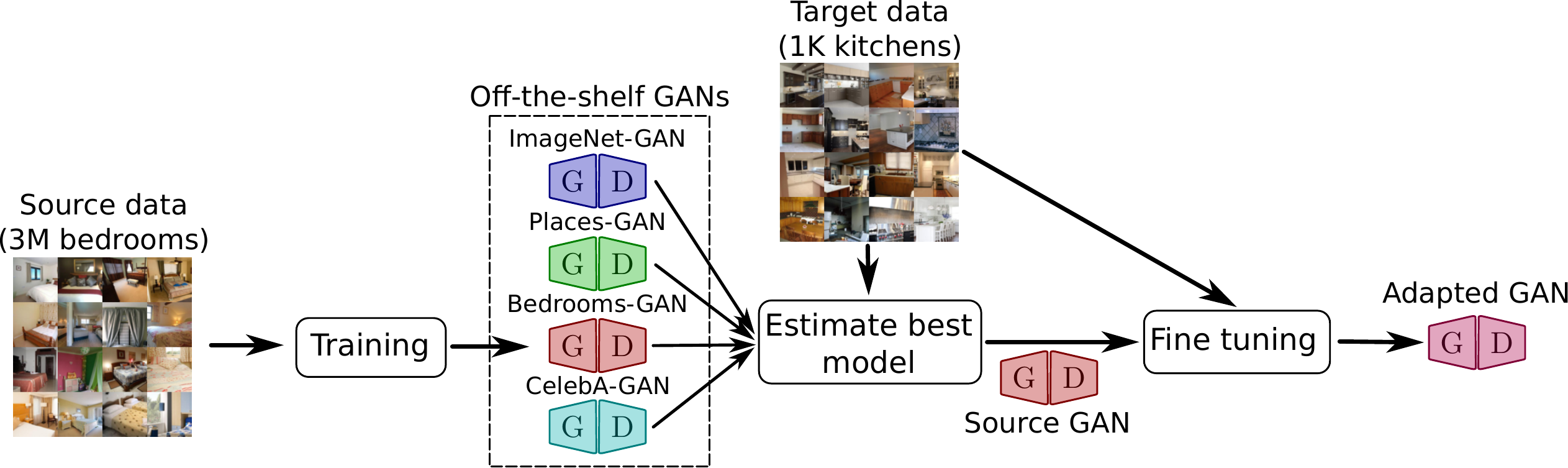}
        \caption{\label{fig:overview} Transferring GANs: training source GANs, estimation of the most suitable pre-trained model and adaptation to the target domain.}
\end{figure}

\subsection{Selecting the pre-trained model}

Selecting a pre-trained model for a discriminative task (e.g. classification) is reduced to simply selecting either ImageNet, for object-centric domains, or Places, for scene-centric ones. The target classifier or fine tuning will simply learn to ignore non-related features and filters of the source network.

However, this simple rule of thumb does not seem to apply so clearly in our GAN transfer setting due to generation being a much more complex task than discrimination. Results in Table~\ref{table:source_target_generated} show that sometimes unrelated datasets may perform better than other apparently more related. The large number of unrelated classes may be an important factor, since narrow yet dense domains also seem to perform better even when they are not so related (e.g. Bedrooms). There are also non-trivial biases in the datasets that may explain this behavior. Therefore, a way to estimate the most suitable model for a given target dataset is desirable, given a collection of pre-trained GANs.

\begin{table}[tb]
	\begin{center}
		\caption{Distance between source generated data $\mathcal{X}^{src}_{gen}$ and target real data $\mathcal{X}^{tgt}_{data}$, and distance between source real $\mathcal{X}^{src}_{data}$ and generated data $\mathcal{X}^{src}_{gen}$.}
		\label{table:distance_source_generated_target_real}
		\begin{tabular}{ccccccc}
			\hline\noalign{\smallskip}
			 & \begin{tabular}[c]{@{}l@{}} Source $\rightarrow$ \\ Target $\downarrow$ \end{tabular} &  ImageNet  & Places & Bedrooms & CelebA \\
			\noalign{\smallskip}
			\hline
			\noalign{\smallskip}
			\multirow{4}{*}{$\textnormal{FID}\left(\mathcal{X}^{src}_{gen},\mathcal{X}^{tgt}_{data}\right)$} & Flowers & \textbf{237.04}& 251.93& 278.80& 284.74 \\
            & Kitchens & 183.27& 180.63& \textbf{70.06}& 254.12 \\
            & LFW & 333.54& 333.38& 329.92& \textbf{151.46} \\
            & Cityscapes & 233.45& \textbf{181.72}& 227.53& 292.66 \\
			\hline
            $\textnormal{FID}\left(\mathcal{X}^{src}_{gen},\mathcal{X}^{src}_{data}\right)$ & Source & 63.46 & 55.66 & 17.30 & 75.84 \\
            \hline
		\end{tabular}
	\end{center}~\label{distance_source_generated_target_real}
    \vspace{-0.9cm}
\end{table}


Perhaps the most simple way is to measure the distance between the source and target domains. We evaluated the FID between the (real) images in the target and the source datasets (results included in the supplementary material). While showing some correlation with the FID of the target generated data, it has the limitation of not considering whether the actual pre-trained model is able or not to accurately sample from the real distribution. A more helpful metric is the distance between the target data and the \textit{generated} samples by the pre-trained model. In this way, the quality of the model is taken into account. We estimate this distance also using FID. In general, there seem to roughly correlate with the final FID results with target generated data (compare Tables~\ref{table:source_target_generated} and \ref{table:distance_source_generated_target_real}).
Nevertheless, it is surprising that Places is estimated as a good source dataset but does not live up to the expectation. The opposite occurs for Bedrooms, which seems to deliver better results than expected. 
This may suggest that density is more important than diversity for a good transferable model, even for apparently unrelated target domains.

In our opinion, the FID between source generated and target real data is a rough indicator of suitability rather than accurate metric. It should taken into account jointly with others factors (e.g. quality of the source model) to decide which model is best for a given target dataset.


\subsection{Visualizing the adaptation process}

One advantage of the image generation setting is that the process of shifting from the source domain towards the target domain can be visualized by sampling images at different iterations, in particular during the initial ones. Figure~\ref{fig:progression_sources} shows some examples of the target domain Kitchens and different source domains (iterations are sampled in a logarithmic scale). 

Trained from scratch, the generated images simply start with noisy patterns that evolve slowly, and after 4000 iterations the model manages to reproduce the global layout and color, but still fails to generate convincing details. Both the GANs pre-trained with Places and ImageNet fail to generate realistic enough source images and often sample from unrelated source classes (see iteration 0). During the initial adaptation steps, the GAN tries to generate kitchen-like patterns by matching and slightly modifying the source pattern, therefore preserving global features such as colors and global layout, at least during a significant number of iterations, then slowly changing them to more realistic ones. Nevertheless, the textures and edges are sharper and more realistic than from scratch. The GAN pre-trained with Bedrooms can already generate very convincing bedrooms, which share a lot of features with kitchens. The larger number of training images in Bedrooms helps to learn transferable fine grained details that other datasets cannot. The adaptation mostly preserves the layout, colors and perspective of the source generated bedroom, and slowly transforms it into kitchens by changing fine grained details, resulting in more convincing images than with the other source datasets. Despite being a completely unrelated domain, CelebA also manages to help in speeding up the learning process by providing useful priors. Different parts such as face, hair and eyes are transformed into different parts of the kitchen. Rather than the face itself, the most predominant feature remaining from the source generated image is the background color and shape, that influences in the layout and colors that the generated kitchens will have.

\section{Transferring to conditional GANs}
\label{sec:cGANtransfer}

Here we study the transferring the representation learned by a pre-trained unconditional GAN to a cGAN~\cite{mirza2014conditional}. 
cGANs allow us to condition the generative model on particular information such as classes, attributes, or even other images.
Let $y$ be a conditioning variable. 
The discriminator $D(x,y)$ aims to distinguish pairs of real data $x$ and $y$ sampled from the joint distribution $p_{data}\left(x,y\right)$ from pairs of generated outputs $G(z,y')$ conditioned on samples $y'$ from $y$'s marginal $p_{data}(y)$. 




\subsection{Conditional GAN adaptation}
For the current study, we adopt the Auxiliary Classifier GAN (AC-GAN) framework of~\cite{odena2016conditional}.
In this formulation, the discriminator has an `auxiliary classifier' that outputs a probability distribution over classes $P(C=y|x)$ conditioned on the input $x$.
The objective function is then composed of the conditional version of the GAN loss $\mathcal{L}_{GAN}$ (eq.~\eqref{eq:GAN}) and the log-likelihood of the correct class.
The final loss functions for generator and discriminator are:
\begin{gather}
\mathcal{L}_{AC-GAN}\left(G\right)=\mathcal{L}_{GAN}\left(G\right)-\alpha_{G}\mathbb{E}\left[\log\left(P\left(C=y'|G(z,y')\right)\right)\right],
\label{eq:L_D}
\\
\mathcal{L}_{AC-GAN}\left(D\right)=\mathcal{L}_{GAN}\left(D\right)-\alpha_{D}\mathbb{E}\left[\log\left(P\left(C=y|x\right)\right)\right],
\label{eq:L_G}
\end{gather}
respectively.  
%
The parameters $\alpha_{G}$ and $\alpha_{D}$ weight the contribution of the auxiliary classifier loss with respect to the GAN loss for the generator and discriminator. In our  implementation, we use Resnet-18~\cite{he2016resnet} for both $G$ and $D$, and the WGAN-GP loss from the equations~\eqref{eq:WGANGP_D} and~\eqref{eq:WGANGP_G} as the GAN loss. Overall, the implementation details (batch size, learning rate) are the same as introduced in section~\ref{GAN_domain_adaptation}. 

In AC-GAN, the conditioning is performed only on the generator by appending the class label to the input noise vector. 
We call this variant `Cond Concat'. We randomly initialize the weights which are connected to the conditioning prior. 
We also used another variant following~\cite{dumoulin2017artistic}, in which the conditioning prior is embedded in the batch normalization layers of the generator (referred to as `Cond BNorm'). In this case, there are different batch normalization parameters for each class. We initialize these parameters by copying the values from the unconditional GAN to all classes.



\subsection{Results}
We use Places~\cite{zhou2014learning} as the source domain and consider all the ten classes of the LSUN dataset~\cite{yu2015construction} as target domain.
We train the AC-GAN with 10K images per class for 25K iterations.
The weights of the conditional GAN can be transferred from the pre-trained unconditional GAN (see section \ref{sec:GAN_basics}) or initialized at random. 
The performance is assessed in terms of the FID score between target domain and generated images.  The FID is computed class-wise, averaging over all classes and also considering the dataset as a whole (class-agnostic case). The classes in the target domain have been generated uniformly. The results are presented in table \ref{table:FID_cGANs}, where we show the performance of the AC-GAN whose weights have been transferred from pre-trained network vs. an AC-GAN initialized randomly. We computed the FID for 250, 2500 and 25000 iterations. At the beginning of the learning process, there is a significant difference between the two cases. The gap is reduced towards the end of the learning process but a significant performance gain still remains for pre-trained networks. We also consider the case with fewer images per class. The results after 25000 iterations for 100 and 1K images per class are provided in the last column of table~\ref{table:classification}.
We can observe how the difference between networks trained from scratch or from pre-trained weights is more significant for smaller sample sizes.
This confirms the trend observed in section~\ref{sec:sizeTarget}: transferring the pre-trained weights is especially advantageous when only limited data is available.

The same behavior can be observed in figure~\ref{fig:user_study} (left) where we compare the performance of the AC-GAN with two unconditional GANs, one pre-trained on the source domain and one trained from scratch, as in section~\ref{sec:transferConf}. The curves correspond to the class-agnostic case (column `All' in the table \ref{table:FID_cGANs}).
From this plot, we can observe three aspects: (i) the two variants of AC-GAN perform similarly 
(for this reason, for the remaining of the experiments we consider only `Cond BNorm');
(ii) the network initialized with pre-trained weights converges faster than the  network trained from scratch, and the overall performance is better; and (iii) AC-GAN performs slightly better than the unconditional GAN.

\setlength{\tabcolsep}{1.5pt}
\begin{table}[t]
	\begin{center}
		\caption{Per-class and overall FID for AC-GAN. Source: Places, target: LSUN}
		\label{table:FID_cGANs}
        \resizebox{\textwidth}{!}{
		\begin{tabular}{cc|cccccccccc|c|c}
\hline
Init & Iter & Bedr & Bridge & Church & Classr & Confer & Dining & Kitchen & Living & Rest & Tower & Avg. & All \\
\hline
\multirow{3}{*}{Scratch}  & 250 &298.4&310.3 &314.4 & 376.6&339.1 & 294.9& 314.2&316.5 &324.4 &301.0 & 319.0 & 352.4\\
                     & 2500 &195.9 &135.0 &133.0 &218.6 &185.3 &173.9 &167.9 &189.3 &159.5 & 125.6& 168.4 & 137.3\\
                     & 25000 & 72.9&78.0 &52.4 & 106.7&76.9 & 40.1&53.9 &56.1 & 74.7 &59.8 & 67.2 &49.6 \\
                     \hline
\multirow{3}{*}{Pre-trained} & 250 &\textbf{168.3} &\textbf{122.1} & \textbf{148.1}&\textbf{145.0} &\textbf{151.6} &\textbf{144.2} &\textbf{156.9} &\textbf{150.1} &\textbf{113.3} &\textbf{129.7} & \textbf{142.9} & \textbf{107.2}\\
                     & 2500 & \textbf{140.8}&\textbf{96.8} &\textbf{77.4} &\textbf{136.0} &\textbf{136.8} & \textbf{84.6}&\textbf{85.5} & \textbf{94.9}&\textbf{77.0} & \textbf{69.4}& \textbf{99.9} & \textbf{74.8}\\
                     & 25000 &\textbf{59.9} &\textbf{68.6} &\textbf{48.2} &\textbf{79.0} & \textbf{68.7}& \textbf{35.2}& \textbf{48.2}& \textbf{47.9}&\textbf{44.4} & \textbf{49.9}& \textbf{55.0} & \textbf{42.7}\\
                     \hline
\end{tabular}
}

\end{center}
\vspace{-0.9cm}
\end{table}

Next, we evaluate the AC-GAN performance on a classification experiment. We train a reference classifier on the 10 classes of LSUN (10K real images per class). Then, we evaluate the quality of each model trained for 25K iterations by generating 10K images per class and measuring the accuracy of the reference classifier for 100, 1K and 10K images per class.
The results show an improvement when using pre-trained models, with
higher accuracy and lower FID in all settings, suggesting that it captures better the real data distribution of the dataset compared to training from scratch.

Finally, we perform a psychophysical experiment with generated images by AC-GAN with LSUN as target. 
Human subjects are presented with two images: pre-trained vs. from scratch (generated from the same condition \texttt{<class>}), and asked `Which of these two images of \texttt{<class>} is more realistic?'
Subjects were also given the option to skip a particular pair should they find very hard to decide for one of them.
We require each subject to provide 100 valid assessments.
We use 10 human subjects which evaluate image pairs for different settings (100, 1K, 10K images per class). The results (Fig.~\ref{fig:user_study} right) clearly show that the images based on pre-trained GANs are considered to be more realistic in the case of 100 and 1K images per class (e.g. pre-trained is preferred in 67\% of cases with 1K images). As expected the difference is smaller for the 10K case.




\setlength{\tabcolsep}{4pt}

\setlength{\tabcolsep}{1.5pt}
\begin{table}[t]
	\begin{center}
		\caption{Accuracy of AC-GAN for the classification task and overall FID for different sizes of the target set (LSUN). 
        }
		\label{table:classification}
        \resizebox{0.9\textwidth}{!}{
		\begin{tabular}{c|c|cccccccccc|c|c}
\hline
\multirow{2}{*}{\#images} & \multirow{2}{*}{Method} & \multicolumn{11}{c}{Accuracy (\%)} & \multirow{2}{*}{FID}\\
 &  & Bedr & Bridge & Church & Classr & Confer & Dining & Kitchen & Living & Rest & Tower & Avg. & \\
\hline
\multirow{2}{*}{100/class}  
& scratch & 23.0 & \textbf{88.2} & \textbf{55.1} & 29.2 & 3.6 & 24.9 & 20.8 & 8.4 & \textbf{89.3} & \textbf{61.6} & 40.4 & 162.9\\
                     & pre-trained & \textbf{35.7} & 72.7 & 45.7 & \textbf{59.4} & \textbf{7.9} & \textbf{38.2} & \textbf{36.3} & \textbf{20.1} & 81.0 & 56.6 & \textbf{45.4} & \textbf{119.1}\\
                     \hline
                     \multirow{2}{*}{1K/class}  & scratch & 49.9 & 78.1 & \textbf{75.1} & 51.8 & 14.6 & 51.2 & 31.2 & 23.2 & \textbf{90.7} & 61.5 & 52.7 & 117.3\\
                     & pre-trained & \textbf{76.4} & \textbf{82.5} & 69.1 & \textbf{80.6} & \textbf{34.2} & \textbf{52.6} & \textbf{62.4} & \textbf{52.9} & 80.5 & \textbf{67.5} & \textbf{65.9} & \textbf{77.5}\\
                     \hline
                      \multirow{2}{*}{10K/class}  & scratch & \textbf{94.9} & 94.3 & 89.6 & 85.0 & 82.4 & \textbf{91.2} & 88.0 & 86.9 & 91.3 & 83.5 & 88.7 & 49.6 \\
                      & pre-trained & 87.1 & \textbf{95.7} & \textbf{90.8} & \textbf{95.1} & \textbf{86.8} & 90.2 & \textbf{88.9} & \textbf{90.1} & \textbf{93.0} & \textbf{88.9} & \textbf{90.8} & \textbf{42.7}\\
                     \hline
\end{tabular}
}
\end{center}
\vspace{-0.9cm}
\end{table}


\begin{figure}
	\centering
    
    \begin{tabu} to 0.98\textwidth {@{} *9{X[c]} *2{X[c]}@{}}  0 \small{(src)} & 2 & 5 & 15 & 39 & 99 & 251 & 632 & 1591 & 4000 & 0 \small{(src)}\\ \end{tabu} \\
	\includegraphics[width=0.98\textwidth]{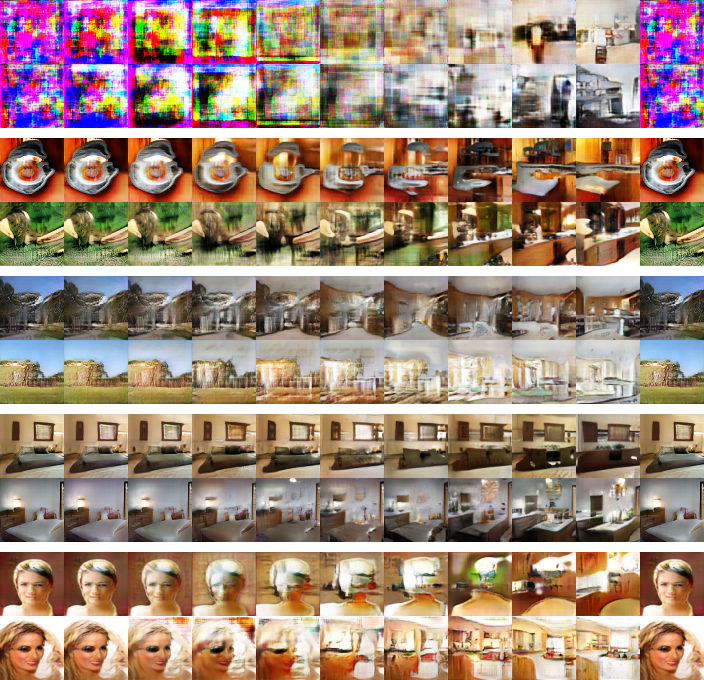}
	\caption{Evolution of generated images (in logarithmic scale) for LSUN kitchens with different source datasets (from top to bottom: from scratch, ImageNet, Places, LSUN bedrooms, CelebA).
    Better viewed in electronic version.}
	\label{fig:progression_sources}
\end{figure}

\begin{figure}
    \centering

   \begin{subfigure}{0.49\textwidth}
   \centering
   \includegraphics[width=\textwidth]{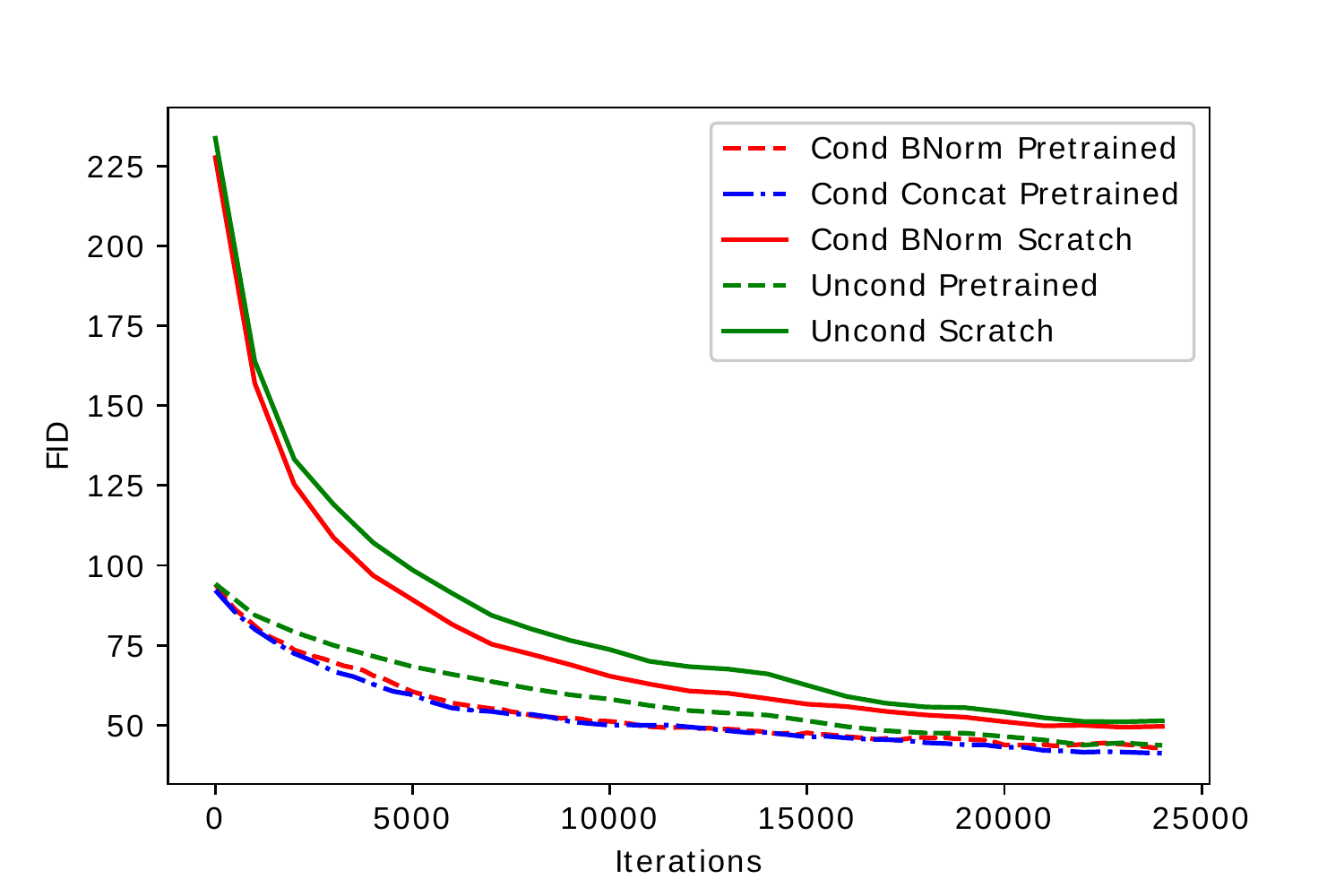}
   \end{subfigure}
   \hspace{5mm}
     \begin{subfigure}{0.37\textwidth }
   \centering
   \includegraphics[width=\textwidth]{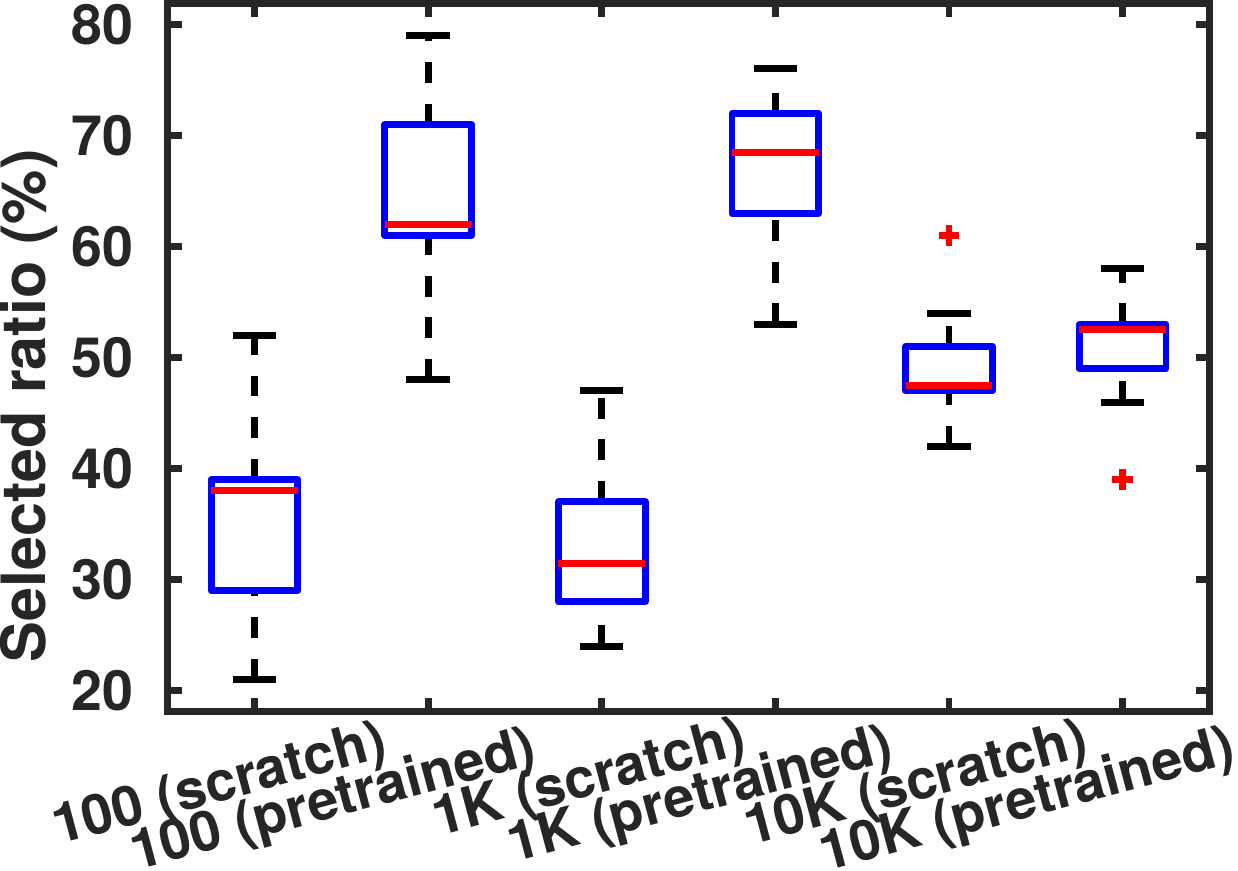}

   \end{subfigure}%
   \caption{(Left) FID score for Conditional and Unconditional GAN (source: Places, target: LSUN 10 classes). (Right) Human evaluation of image quality.}
   \label{fig:user_study}
\end{figure}

\section{Conclusions}
We show how the principles of transfer learning can be applied to generative features for image generation with GANs. GANs, and conditional GANs, benefit from transferring pre-trained models, resulting in lower FID scores and more recognizable images with less training data. Somewhat contrary to intuition, our experiments show that transferring the discriminator is much more critical than the generator (yet transferring both networks is best). However, there are also other important differences with the discriminative scenario. Notably, it seems that a much higher density (images per class) is required to learn good transferable features for image generation, than for image discrimination (where diversity seems more critical). As a consequence, ImageNet and Places, while producing excellent transferable features for discrimination, seem not dense enough for generation, and LSUN data seems to be a better choice despite its limited diversity. Nevertheless, poor transferability may be also related to the limitations of current GAN techniques, and better ones could also lead to better transferability.

Our experiments evaluate GANs in settings rarely explored in previous works and show that there are many open problems. These settings include GANs and evaluation metrics in the very limited data regime, better mechanisms to estimate the most suitable pre-trained model for a given target dataset, and the design of better pre-trained GAN models.

\minisection{Acknowledgement}
Y. Wang and C. Wu acknowledge the Chinese Scholarship Council (CSC) grant
No.201507040048 and No.201709110103. L. Herranz acknowledges the
European Union research and innovation program under the Marie
Skłodowska-Curie grant agreement No. 6655919. This work was supported
by TIN2016-79717-R, and the CHISTERA project M2CR
(PCIN-2015-251) of the Spanish Ministry, the CERCA
Program of the \emph{Generalitat de Catalunya}. We also acknowledge the generous GPU
support from NVIDIA.

\bibliographystyle{splncs}
\bibliography{shortstrings,eccv}

\fi



\iftrue

\appendix
\section*{Supplementary Material}
\section{Distances between source and target data}
\setcounter{table}{0}
\renewcommand{\thetable}{A\arabic{table}}
Table~\ref{table:distance_source_real_target_real} shows the FID between the real images in the source and target datasets, which could be used as an estimation of which pre-trained GAN (on a source dataset) may be a good choice to adapt to a particular target dataset. In most of the cases, the lowest value in Table~\ref{table:distance_source_real_target_real} also corresponds to the lowest value in Table~1.

\setlength{\tabcolsep}{4pt}
\begin{table}
{\small
	\begin{center}
		\caption{Distance between source real data and target real data.}
		\label{table:distance_source_real_target_real}
		\begin{tabular}{ccccccc}
			\hline\noalign{\smallskip}
			Distance & \begin{tabular}[c]{@{}l@{}}Source $\rightarrow$ \\ Target $\downarrow$ \end{tabular} &  ImageNet  & Places & Bedrooms & CelebA \\
			\noalign{\smallskip}
			\hline
			\noalign{\smallskip}
			\multirow{4}{*}{$\textnormal{FID}\left(\mathcal{X}^{src}_{data},\mathcal{X}^{tgt}_{data}\right)$} & Flowers & \textbf{187.52} & 292.36 & 270.09 & 317.21 \\
            & Kitchens & 139.81 & 99.88 & \textbf{66.54} & 311.06 \\
            & LFW & 266.50 & 326.76 & 318.98 & \textbf{44.12} \\
            & Cityscapes & 205.04 & \textbf{143.55} & 221.65 & 349.28\\
            \hline
		\end{tabular}
	\end{center}
    }
    \vspace{-0.9cm}
\end{table}





\begin{figure}[b]
	\centering
	\includegraphics[width=0.4\textwidth]{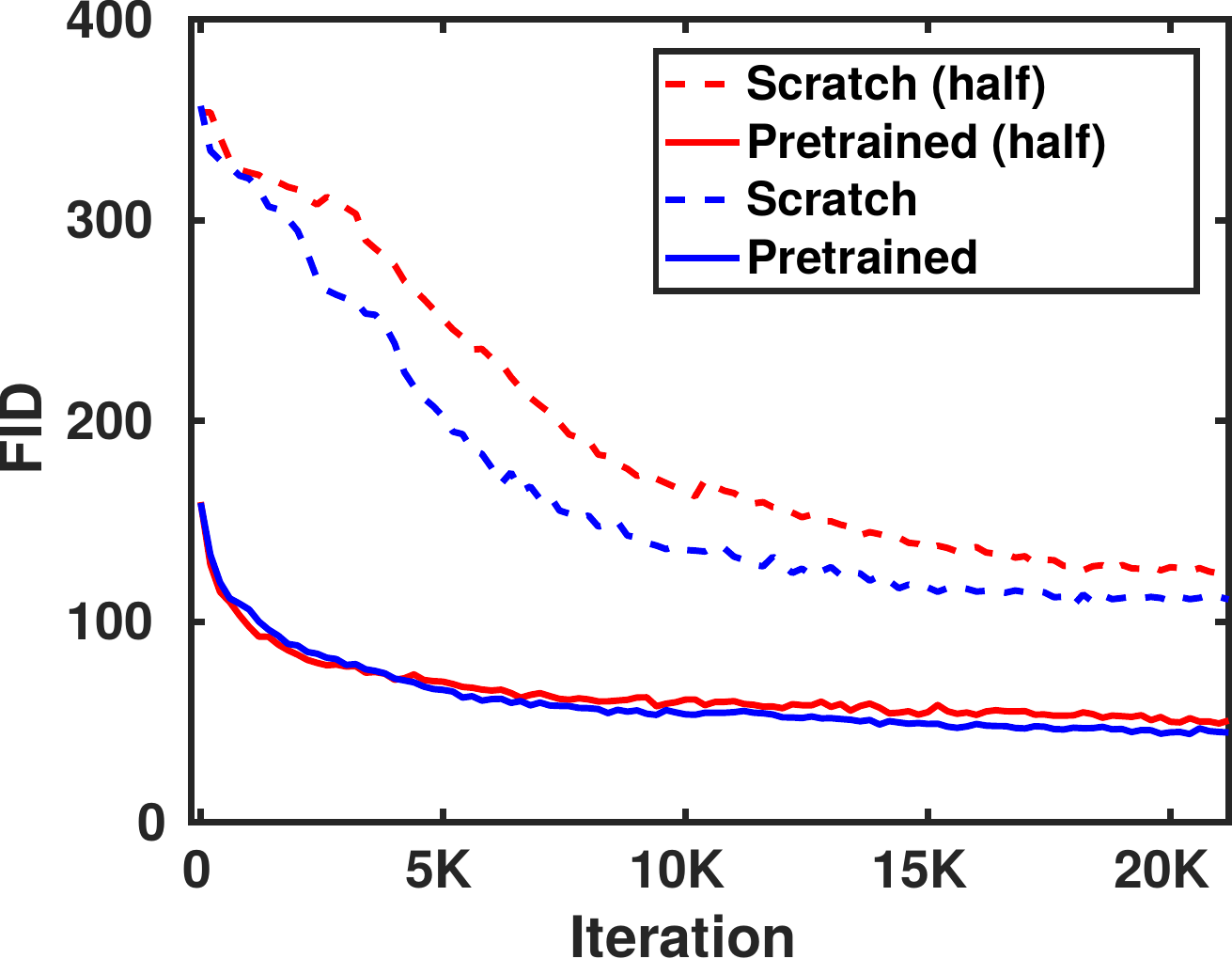}
	\caption{Model capacity }
	\label{fig:model_capacity}
\end{figure}

\section{Model capacity}
In order to check how important the capacity of the network is for transferring GAN features, we performed an additional experiment where we reduced the capacity of the network to half. We trained a source GAN with ImageNet, but in this case we reduced the number of filters in each layer to half its original value (with respect to the architecture used throughout our paper, from WGAN-GP~[25]). The model is then fine tuned with 10K images from LSUN Bedrooms. The results shown in Fig.~\ref{fig:model_capacity} suggest that also a lower capacity GAN adapting pre-trained features obtains significantly better results.

\section{Images sampled from the models}
We also show examples of images sampled from each of the source models after fine tuning 5K iterations with Flowers (Fig.~\ref{fig:supp_flowers}), Kitchens (Fig.~\ref{fig:supp_kitchens}), LFW (Fig.~\ref{fig:supp_lfw}), and cityscapes (Fig.~\ref{fig:supp_cityscapes}).

\begin{figure}[tb]
	\centering
	\includegraphics[width=\textwidth]{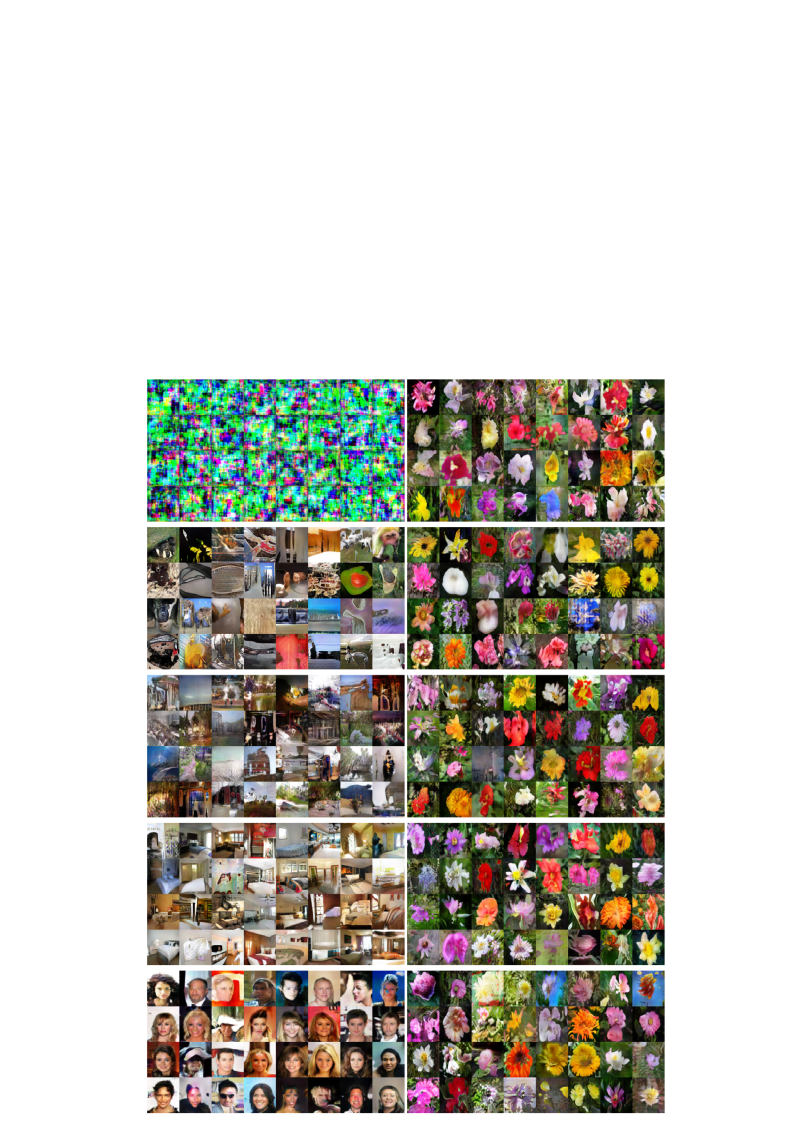}
	\caption{Images sampled from each of the source models (left) and after fine tuning 5K iterations with Flowers (right). From top to bottom: from scratch, ImageNet, Places, LSUN bedrooms, CelebA.}
	\label{fig:supp_flowers}
\end{figure}

\begin{figure}
	\centering
	\includegraphics[width=\textwidth]{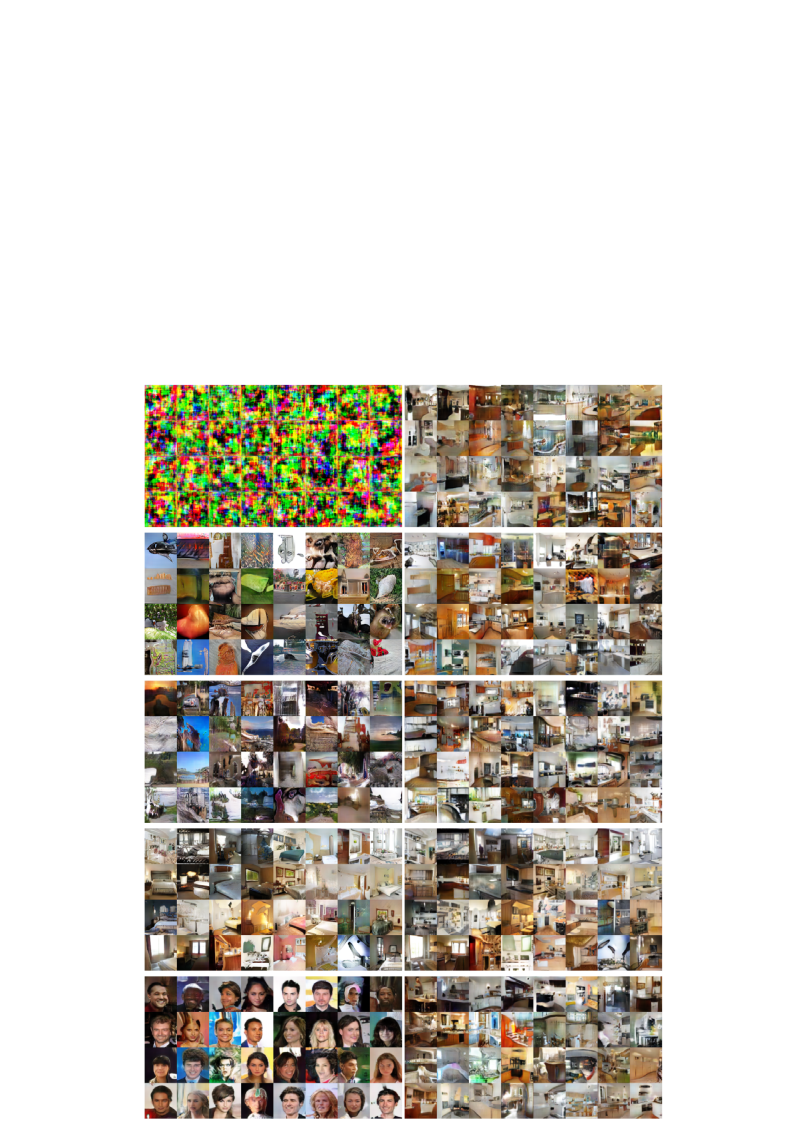}
	\caption{Images sampled from each of the source models (left) and after fine tuning 5K iterations with Kitchens (right). From top to bottom: from scratch, ImageNet, Places, LSUN bedrooms, CelebA.}
	\label{fig:supp_kitchens}
\end{figure}

\begin{figure}
	\centering
	\includegraphics[width=\textwidth]{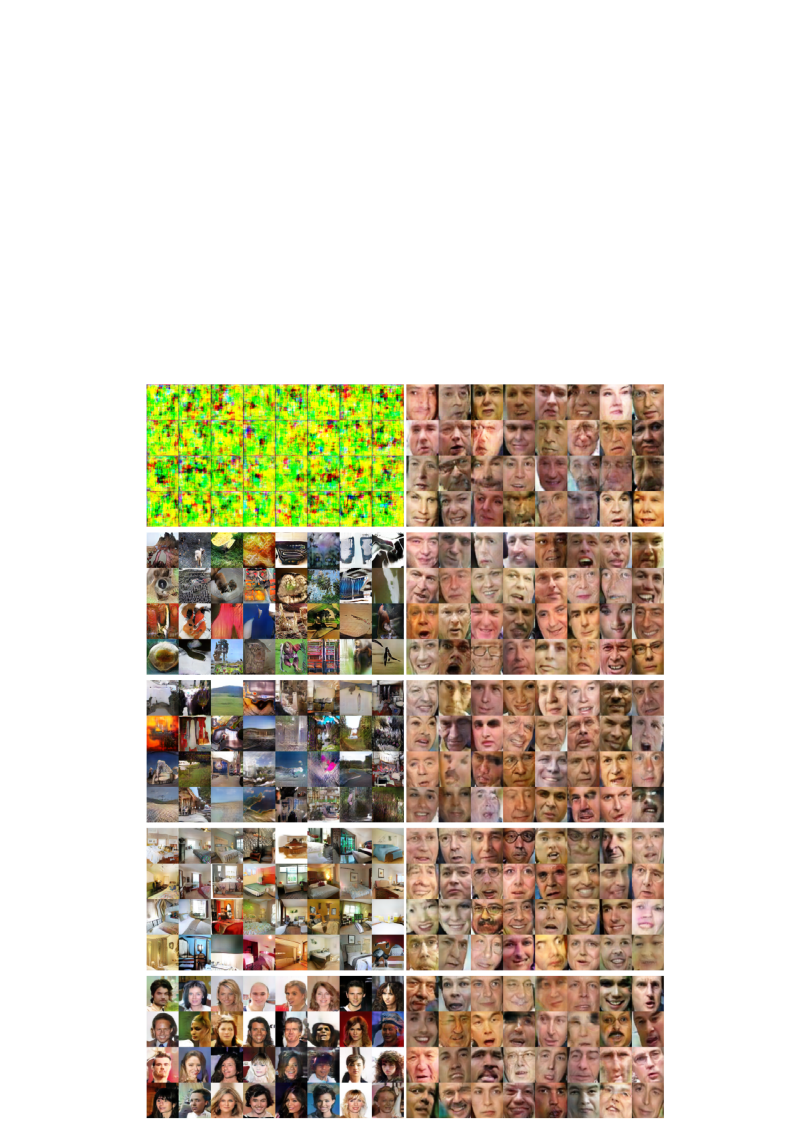}
	\caption{Images sampled from each of the source models (left) and after fine tuning 5K iterations with LFW (right). From top to bottom: from scratch, ImageNet, Places, LSUN bedrooms, CelebA.}
	\label{fig:supp_lfw}
\end{figure}

\begin{figure}
	\centering
	\includegraphics[width=\textwidth]{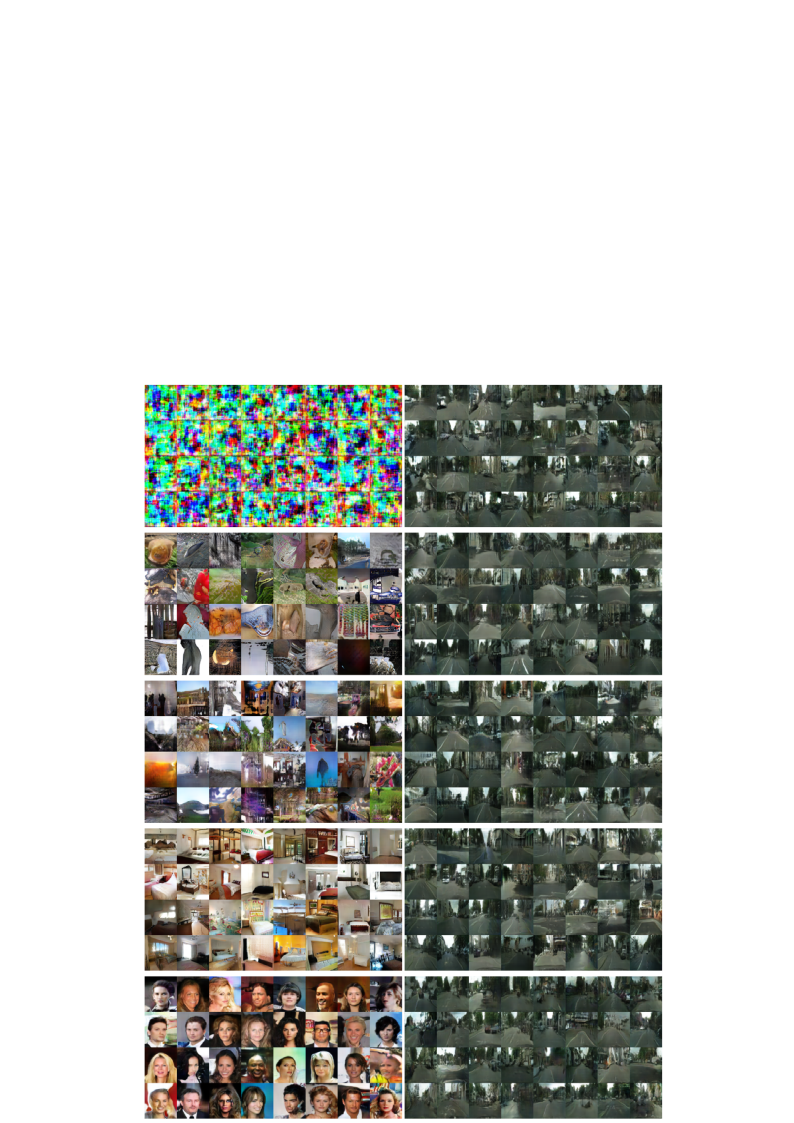}
	\caption{Images sampled from each of the source models (left) and after fine tuning 5K iterations with Cityscapes (right). From top to bottom: from scratch, ImageNet, Places, LSUN bedrooms, CelebA.}
	\label{fig:supp_cityscapes}
\end{figure}

\fi
\end{document}